\documentclass[utf8, a4paper]{article} %

\usepackage{natbib}
\usepackage{tabularx}
\usepackage{hyperref}
\usepackage{graphicx}
\usepackage[affil-it]{authblk}

\title{RadEx: A Framework for Structured Information Extraction from Radiology Reports based on Large Language Models} 

\author[1,$\star$]{Daniel Reichenpfader}
\author[2]{Jonas Knupp}
\author[2]{André Sander}
\author[1]{Kerstin Denecke}

\affil[1]{Institute for Patient-centered Digital Health, Bern University of Applied Sciences, Biel, Switzerland}
\affil[2]{ID Suisse AG, St. Gallen, Switzerland}
\affil[$\star$]{Corresponding author: Daniel Reichenpfader, daniel.reichenpfader@bfh.ch}

\date{\today}

\begin{document}

\maketitle

\begin{abstract}
Annually and globally, over three billion radiography examinations and computer tomography scans result in mostly unstructured radiology reports containing free text. Despite the potential benefits of structured reporting, its adoption is limited by factors such as established processes, resource constraints and potential loss of information. However, structured information would be necessary for various use cases, including automatic analysis, clinical trial matching, and prediction of health outcomes. This study introduces RadEx, an end-to-end framework comprising 15 software components and ten artifacts to develop systems that perform automated information extraction from radiology reports. It covers the complete process from annotating training data to extracting information by offering a consistent generic information model and setting boundaries for model development. Specifically, RadEx allows clinicians to define relevant information for clinical domains (e.g., mammography) and to create report templates. The framework supports both generative and encoder-only models and the decoupling of information extraction from template filling enables independent model improvements. Developing information extraction systems according to the RadEx framework facilitates implementation and maintenance as components are easily exchangeable, while standardized artifacts ensure interoperability between components.  
\end{abstract}

\newpage
\section{Introduction}

More than three billion conventional radiography (X-Ray) examinations and CT scans are performed in medical practice, worldwide, each year \citep{maheshPatientExposureRadiologic2023}. These diagnostic procedures enable physicians to detect and diagnose various diseases and decide on appropriate treatment options. The radiologist's interpretations of the acquired images are described in radiology reports. These reports are usually dictated freely and therefore consist of unstructured text. Although there are approaches toward implementation of semi- or fully-structured radiology reports, these aspirations often fail due to resistance against changing established processes and time and resource constraints. Furthermore, clinicians resist additional steps in interacting with a system \citep{poolStructuredReportingRadiology2022, rochaEvidenceBenefitsAdvantages2020}. Beyond, there is still little evidence that radiology can benefit from structured reporting \citep{nobel2022structured}. 

However, Natural Language Processing (NLP) methods can be used to automatically extract clinically relevant information contained within these free-text reports. Extracting and standardizing information would not only facilitate quality management of radiological reports, but makes existing databases of unstructured reports accessible for clinical research without loosing details for the diagnostic process. Going one step further, the extracted information can be mapped to a predefined set of values and therefore standardized based on reporting and data systems (RADS). For example, the BI-RADS framework specifies the information to be presented in mammography reports as well as associated value sets \citep{radiology2013acr}. Similar manuals exist for other clinical domains like liver and lung examinations. Reporting based on these guidelines, however, is not compulsory in many hospitals, leading to missing values when trying to extract such information automatically.

Several research projects have been conducted to develop systems that extract information from radiology reports based on machine learning (ML) and deep learning (DL) methods. Starting in 2018, a new generation of DL-architectures, namely Large Language Models (LLM), revolutionized the NLP sector and its applications in medicine \citep{thirunavukarasuLargeLanguageModels2023}. Although LLMs brought a significant improvement in various NLP tasks, evidence on their application to extracting information from radiology-reports is still limited. A first scoping review is currently in progress \citep{reichenpfaderLargeLanguageModelbased2023}). Although traditional approaches, e.g., rule-based systems, become increasingly specialized and specified, newer DL-based approaches might be superb due to latest technological improvements. However, the latter require in turn a high amount of manually labelled or annotated training data; data which is especially in the medical domain difficult to acquire due to the sensitive nature of medical data and specific privacy regulations that therefore need to be adhered to \citep{mostertBigDataMedical2016}. Current approaches are often limited regarding the number of different concepts extracted or its application to a narrow domain \citep{smitCombiningAutomaticLabelers2020, kulingBIRADSBERTUsing2022, woodAutomatedLabellingUsing2020, yangClinicalConceptExtraction2020}.

With RadEx, we herewith describe the architecture of an end-to-end framework for automated structuring of radiology reports. The framework is agnostic regarding the model implementation, since models might be exchanged without major adaptions to other components. Other than choice of model, various other challenges are rising when designing an end-to-end framework for structuring of radiology reports: These challenges include correct management of source data, management of the annotation process, defining an information model for extraction, architecture planning and implementation, interfaces and processes, and foremost the definition what is considered to be clinically relevant. What follows is a description of the framework's components and artifacts. Subsequently, a concrete implementation is described. Finally,the  limitations of the framework as well as its potential and further research directions are discussed.

\section{Framework}

\begin{figure*}[htbp]
\centering
\includegraphics[scale=0.5]{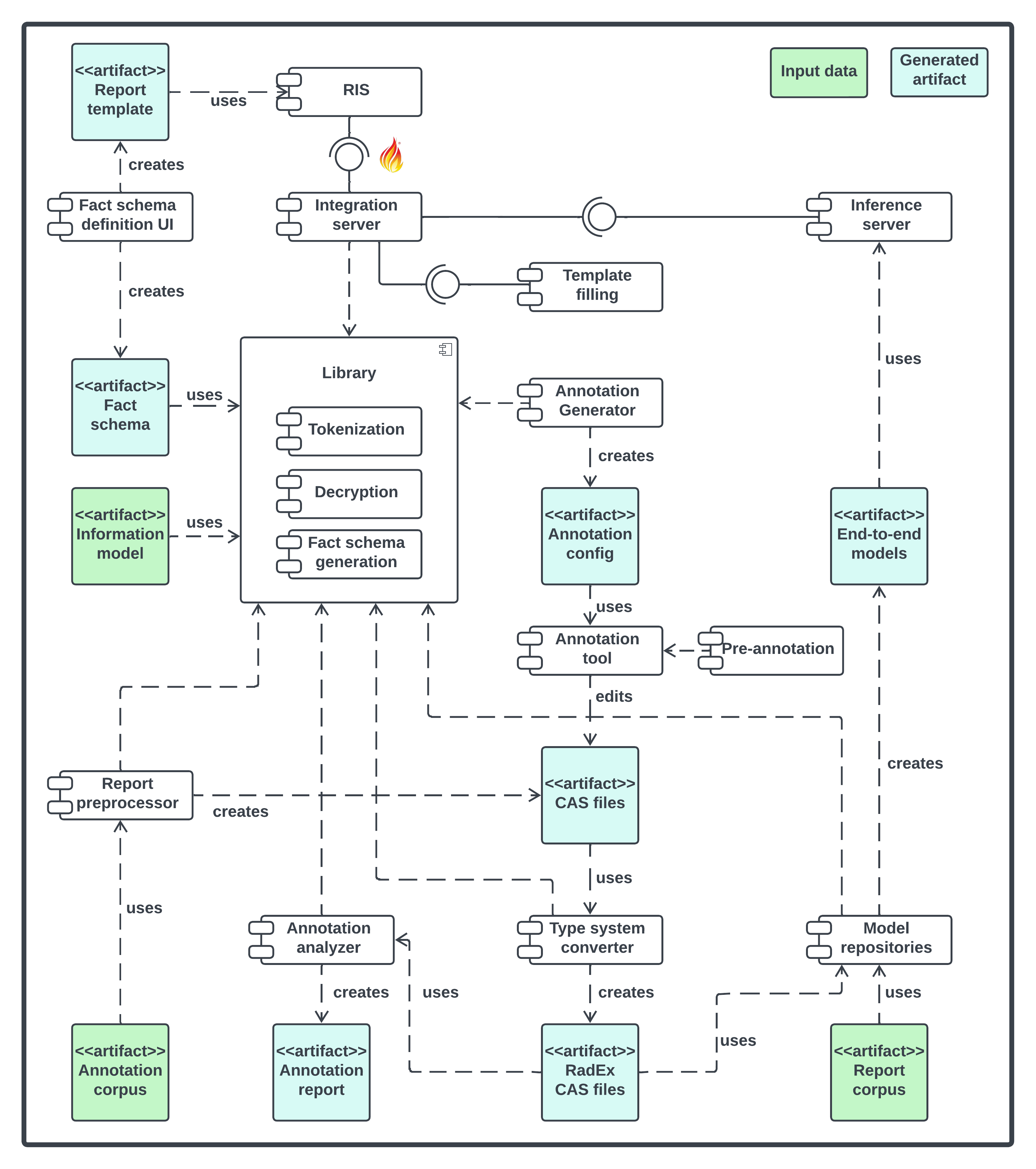} 
\caption{UML component diagram of the RadEx framework. Abbreviations: CAS = Common Analysis Structure, RIS = Radiology Information System, UI = User Interface.}
\label{fig:system-architecture}

\end{figure*}

In the following, we describe the detailed RadEx framework as an assembly of artifacts and components, as depicted in \autoref{fig:system-architecture}. According to the UML standard, a component is a “modular unit with well-defined interfaces”, while an artifact is considered as an “item of information used or produced by a software development process” \citep{objectmanagementgroupUnifiedModelingLanguage2017}. These ten artifacts and 15 components enable the development of systems for automated structuring of clinical texts.

\subsection{Artifacts}

In the following sub-chapters, the ten artifacts are described in detail. Each artifact either represents an exchange format between components or information needed to instantiate a component.

\subsubsection{Information model}
\label{sec:information-model}
 The RadEx \emph{Information model} provides the basic boundary conditions for information to be extracted from radiology reports. It is based on the model proposed by \citet{steinkampCompleteStructuredInformation2019b} and consists of facts, anchor entities and modifiers. A fact is a clinical assertion that is manifested as a continuous text span in the report. Every fact contains exactly one anchor entity, which is the central word or phrase within the fact. Optional modifiers provide additional information about a fact (and consequently also about the anchor entity).
 
For formal specification of the annotation process, the RadEx information model is defined as a UIMA type system file (XML). UIMA (Unstructured Information Management Architecture) is an open-source framework and software infrastructure that supports the development, discovery, composition, and deployment of NLP applications. UIMA is part of the Apache Software Foundation \citep{OASIS:UIMA:2009}.

\subsubsection{Fact schema}

\begin{figure}[!ht]
\centering
\includegraphics[scale=0.5]{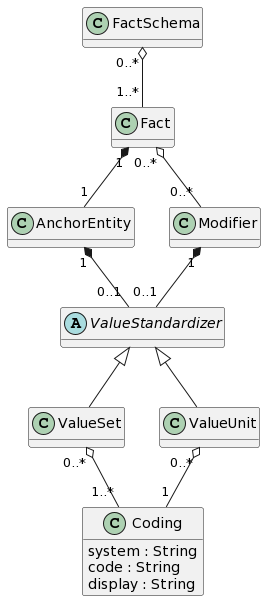} 
\caption{UML class diagram of the RadEx fact schema}
\label{fig:information-model}

\end{figure}

We extend the information model of \citet{steinkampCompleteStructuredInformation2019b} to include value standardizers to enable template filling. Template filling is the process of assigning the extracted text spans to one or more predefined values or, in case of numerical data (length, time, date), standardizing units. 

Therefore, value standardizers are implemented as either value sets or value units: Value sets describe which value(s) a modifier or anchor entity is expected to be mapped to. By defining cardinality properties in value sets, it is possible to model single- and multiple-choice mappings. Value units, however, define the standardized target unit of a modifier, based on an appropriate code system (e.g., Unified Code for Units of Measure (UCUM)). For the example provided in \autoref{tab.fact.ex}, a value set could define the SNOMED Clinical Terms (SNOMED-CT) codings “7771000 | Left”, “24028007 | Right” and “261665006 | Unknown” as possible values for the modifier laterality.

\begin{table}[!h]
\begin{center}
\begin{tabularx}{\columnwidth}{|X|X|X|}
\hline
      \textbf{Schema element} & \textbf{Instance} & \textbf{Example}\\
      \hline
      Fact & Mass \newline described & \textit{No suspicious focal findings distinguishable on the right side}\\
      \hline
      Anchor entity & Mass & \textit{focal findings}\\
      \hline
     Modifier & Negation & \textit{no} \\
      \hline
        Modifier & Dignity & \textit{suspicious} \\
      \hline
      Modifier & Laterality & \textit{on the right side} \\
      \hline
\end{tabularx}
\caption{Example of a fact schema instance and a corresponding sample text extracted from a radiology report.}
\label{tab.fact.ex}
 \end{center}
\end{table}

The total of all defined facts for one use case, including their corresponding anchor entity and modifier definitions as well as value standardizers, is called a \emph{Fact schema} and specified in an application-defined JSON object. Following an object-oriented programming paradigm, a \emph{Fact schema} is an instance of the \hyperref[sec:information-model]{\emph{Information model} artifact}, augmented by value standardizers.

\subsubsection{Report template}
A \emph{Report template} contains a subset of facts and their value standardizers defined in a fact schema. It is used to determine which facts should be extracted from a report and which values the entities are mapped to. 

\subsubsection{Annotation configuration}

The artifact \textit{Annotation configuration} is a JSON file that defines the proprietary annotation structure for a specific implementation of the component \emph{Annotation tool}. The configuration is automatically created from a defined \emph{Fact schema} and used to set up an annotation project.

\subsubsection{Annotation report}
The \emph{Annotation report} is a JSON file that contains the aggregated and individual Inter-Annotator-Agreement (IAA) scores for facts, anchor entities, and modifiers. It is generated by the component \emph{annotation analyzer}. 

\subsubsection{CAS files and RadEx CAS files}
CAS (Common Analysis Structure) is a data structure used to represent the input data and the annotations generated by annotators. It provides a uniform way to store and manipulate data and conforms to a specific UIMA type system \citep{UIMA:CAS:2004}.
\emph{CAS files} are generated by the annotation tool and contain the report texts sampled from the annotation corpus and their annotations. These framework-specific \emph{CAS files} are translated by the \emph{Type system converter} component into \emph{RadEx CAS files} implementing the RadEx \emph{Information model}. 

\subsubsection{Annotation and report corpus}

Both the annotation and the report corpus are used to train information extraction models. While the annotation corpus is used to create annotated reports used in supervised ML approaches, the report corpus is intended to contain a larger number of files used for unsupervised ML approaches. The report data are encrypted and only decrypted upon use for model training and/or validation. In order not avoid performing any changes on source data, corpora should be persisted in a database and accessed programmatically. 

\subsubsection{End-to-End models}

\emph{End-to-end model} artifacts contain all information needed for creating a serialized model to served by the \emph{Inference server} component. These data include a.o., framework-specific configuration information, the model vocabulary, tokenizer-related information as well as the model weights. 

\subsection{Components}

In the following sections, we provide a detailed description of the components depicted in \autoref{fig:system-architecture}. Components are instantiated by artifacts and produce or use them to exchange and persist information.

\subsubsection{Integration server}

The \emph{Integration server} component coordinates the processing of an incoming report to be analyzed. It exposes a FHIR API endpoint to the \emph{RIS} component. This endpoint accepts a report text as well as a report template to be filled. Upon inference, the server returns the populated template to the RIS component. A report template is represented as FHIR Questionnaire resource. The populated template is returned as FHIR QuestionnaireResponse resource to the \emph{Radiology Information System (RIS)}.

\subsubsection{Template filling}

The \emph{Template filling} component applies either ML- or heuristic-based NLP methods to the extracted information. While the models served by the inference server are intended to extract and label information according to the fact schema, the template filling algorithms map the extracted information to the defined value standardizers defined in the fact schema and therefore report template.  

\subsubsection{Inference server}

The \emph{Inference server} component provides inference for a certain trained end-to-end model pipeline to the \emph{Integration server}. As each pipeline consists of different steps that each might be executed in a different environment (CPU/GPU), an individual inference endpoint is implemented per model, facilitated by a model serving framework. 

\subsubsection{Fact schema definition user interface (UI)}
This component provides a user interface to define a fact schema which is structurally based on the \emph{RadEx information model}. Based on a fact schema, subsets thereof are designed, which we refer to as report templates. A report template ultimately defines the resulting structure of the information extraction pipeline.

\subsubsection{Annotation generator}
The \emph{Annotation generator} takes the fact schema provided by the library component and converts it to a layer specification file to be imported into the annotation tool component. Depending on the annotation tool used, different adapters are needed for conversion. 

\subsubsection{Library}

Functionalities that are used by several RadEx components are bundled into the \emph{Library} component. Its subcomponents include centralized tokenization, data decryption and fact schema provision. 

Centralized tokenization ensures the same tokenization process for training data and upon inference. Data decryption is needed for data-preserving model training on infrastructure hosted by third parties. Training data is symmetrically encrypted and stored on the system. It is decrypted only during the training procedure, providing the key as a command-line argument.  The library contains helper functions to retrieve the current fact schema and its details, including relations between facts, tokens, and modifiers. The library itself retrieves the act schema definition from the \emph{Fact schema definition UI} component.

\subsubsection{Annotation tool}

The \emph{Annotation tool} provides a user interface for manual annotation of clinical reports by clinicians. The annotation tool is set up with an annotation configuration produced by the \emph{Annotation generator} component. 

The annotation tool must provide functionalities to manage annotators, workload assignment and annotation curation. Furthermore, the component supports integration of pre-annotation components, as described below. 

\subsubsection{Pre-annotation}
In addition to the manual annotation of reports, the RadEx framework also supports the development and application of \emph{Pre-annotator} components. These components are integrated into the annotation tool component. Such a tool might apply basic rule- or heuristic-based approaches in order to reduce annotation effort. For example, clinicians might annotate recurring phrases once and then apply these annotations to the whole corpus of annotation documents (phrase-based annotation). Another option is to use the defined fact schema to apply simple rule-based approaches to find and automatically annotate simple text spans. 

\subsubsection{Report pre-processor}
The \emph{Report pre-processor} is responsible for the pre-processing of reports before being imported in the annotation tool. The component supports fixing of encoding errors, removal of duplicates, tokenization, stratified sampling based on corpus classes, calculation of token statistics as well as transforming the source corpus into the format required for the annotation tool (CAS XML files). 

\subsubsection{Annotation analyzer}
The \emph{Annotation analyzer} generates a report based on files annotated using the annotation tool and converted to the RadEx type system. The analyzer provides granular scores of annotated entities, which are used to gauge annotation quality and manage the annotation process. 

\subsubsection{Type system converter}
This component serves as an adapter to convert annotated files between the (proprietary) type system of the annotation tool implementation and the predefined RadEx CAS type system. Using this adapter, model development is decoupled from the annotation tool. 

\subsubsection{Model repositories}
A \emph{Model repository} includes the source code to train an end-to-end model based on the annotated and converted RadEx CAS documents, as well as a report corpus used for self-supervised training. An end-to-end model pipeline deployed on the \emph{Inference server} component expects a radiology report text and returns a list of annotated text spans (including facts, anchors entities and modifiers). Depending on the specific model architecture, the model source code might define various tasks, including tokenizer training, pre-training, further-pretraining, fine-tuning, instruction-tuning, etc. A specific repository structure must be respected, including tools to ensure code quality. For training, encrypted reports are stored within the model repository. The encryption key is provided as a command-line argument. Upon completion of training of a model, an end-to-end model artifact is generated, which comprises all necessary data for deploying the model on the inference server component. Model repositories furthermore include data pre-and post-processors, scripts for data validation, logging and visualization utilities, hyperparameter tuning and optimization scripts as well as means for model evaluation. 

\subsubsection{Radiology Information System (RIS)}

We envision a future RadEx framework integrated into existing RIS solutions: A report template created by the \emph{Fact schema definition UI} component is chosen and sent to the integration server  together with the report text to be analyzed  via a FHIR-based interface.  

\section{Case study}

To show the feasibility of the proposed framework, a system prototype was implemented. The case study focused on extracting information from mammography reports. First, a fact schema was iteratively developed by clinicians and medical engineers, comprising 24 facts, 24 corresponding anchors, and 66 modifiers, based on the BI-RADS quality assurance tool \citep{radiology2013acr}. These facts are intended to represent all relevant information contained in a mammography report.

Next, a stratified sample of 210 mammography reports was annotated according to the fact schema. As annotation tool, the open-source software Inception was installed locally \citep{klieINCEpTIONPlatformMachineAssisted2018}. We then used these annotated reports to fine-tune medBERT.de, an LLM based on the original BERT architecture and pre-trained on 4.7 Million German medical documents, comprising medical texts, clinical notes, research papers, and healthcare-related documents \citep{bressemMEDBERTComprehensiveGerman2023}.
The model repository was implemented based on several open-source NLP libraries maintained by HuggingFace \citep{wolfHuggingFaceTransformersStateoftheart2020} and PyTorch \citep{paszkePyTorchImperativeStyle2019}. The inference server is based on BentoML, an open-source AI application framework \citep{yangBentoMLFrameworkBuilding2024}. In the online repository (see data availability statement), we make selected artifacts available, including the UIMA type system, the annotation configuration to be imported in Inception, as well as examples for Inception CAS and RadEx CAS files. These files are based on a mammogram screening example provided by the National Electrical Manufacturers Association in the explanatory information to the DICOM standard \citep{nationalelectricalmanufacturersassociationDICOMPS3172013}. 

The final implementation comprises an extractive question answering pipeline to extract all facts contained within a report, followed by a sequence labelling pipeline to label all tokens contained within each fact. The question answering pipeline achieves an average F1 score of 90,68 \% (metric: squad\_v2) and the sequence labelling pipeline achieves an average F1 score of 79 \% (metric: seqeval). Components not implemented in the system prototype comprise the fact schema definition UI, template filling, integration server, and RIS, see \autoref{fig:system-architecture}. Furthermore, value standardizers were not defined. However, the implementation of the full framework, as well as ablation studies for model improvement, are currently being conducted. The results are published in a separate paper. 

\section{Discussion}

In this paper, we describe the architecture of an end-to-end framework for realizing radiological information extraction projects. From a clinical perspective, a system based on the RadEx framework provides the following functionalities: 
\begin{itemize}
    \item Definition of relevant clinical information to be extracted (so-called “facts”) for a specific domain (e.g., mammography), including the definition of expected values or value sets
    \item Execution of an annotation process required for model development 
    \item Definition of a report template, representing a subset of facts and associated value sets
    \item Automated completion of a report template based on an input report
\end{itemize}

From a technical perspective, a system based on the RadEx framework provides the following functionalities: 

\begin{itemize}
    \item Provision of a generic information model (UIMA type system) ensuring consistency among annotation, model development and inference
    \item Management of an annotation process
    \item Boundary conditions for the development of models used for information extraction and template filling
    \item Boundary conditions for model deployment and inference
    \item Boundary conditions for internal and external interfaces
\end{itemize}

As the definition of clinically relevant information and their corresponding expected values is embedded in the framework architecture and processes, we offer a flexible framework regarding extraction targets as well as models, which distinguishes our work from other approaches. Below, we describe open research questions as well as strengths and limitations of our framework.

The information model embedded in the RadEx framework might streamline annotation processes and model development due to providing a “single source of truth”. However, estimating the time required to carry out a completely new annotation project is a challenge. The  required resources depend not only on the number and granularity of clinical information to be extracted (and therefore also annotated), but are also technically influenced by the choice of model architecture. Rule-based models typically demand significant effort for setting up the heuristics used for extraction, whereas DL-based approaches learn these heuristics automatically.  However, for rule-based systems, labeling might be limited to annotating documents for model evaluation only, DL-based models depend on a large amount of annotated training data. Looking into the future, the application of LLMs could again reduce the required labeling effort due to zero-, single- or few-shot learning. Applying these methods, (almost) all information needed for information extraction is already embedded in a LLM. 

There are several possible adaptions to the information model presented in this paper that might impact model performance. For example, modifiers (entities that further describe a fact) shared by two or more facts might be merged. Similarly, facts that share the same set of modifiers might be merged. While merging of entities increases the amount of available labeled training data per fact or anchor and therefore potentially increasing model performance, the subsequent template filling might be affected due to increased data heterogeneity. In addition, the complete or partial transferability of a fact scheme to another institution remains unclear; model performance will depend on the degree of difference between the report wording.

As shown in \autoref{fig:system-architecture}, our framework is based on a total of ten artifacts, required to realize an implementation of the RadEx framework. These artifacts might be partially replaced by the establishment of interfaces between some components. For example, a \emph{Report template} might be stored in and accessed directly from the \emph{Fact definition} component. Furthermore, the ongoing improvement and monitoring of models might be automatized according to the Machine Learning Operations (MLOps) paradigm, as outlined by \citet{kreuzbergerMachineLearningOperations2023}. The authors describe principles, components, roles, and workflows of automating and operationalizing ML-based systems. 

RadEx shows several similarities with GATE, an open-source platform for text processing, which was last updated in 2009 \citep{cunninghamGettingMoreOut2013}: Both frameworks rely on decoupled and replaceable software components, making them easily extensible and maintainable. Due to its age, GATE offers a large developer community for support. The in-build IE component of GATE called ANNIE, however, is based on finite state algorithms and the JAPE language as opposed to the RadEx framework which was developed for DL-based IE models. It remains unclear whether GATE's architecture is flexible enough to also integrate state-of-the-art, LLM-based IE models.

A similar framework called DeepNLPF was presented at LREC 2020. According to the authors, it “promotes an easy integration of third-party NLP tools
allowing the preprocessing of natural language texts at
lexical, syntactic, and semantic levels" \citep{rodriguesDeepNLPFFrameworkIntegrating2020}. However, DeepNLPF only corresponds to the \emph{model repository} component, and could be therefore integrated itself into RadEx. Other approaches, like the one described by \citet{kocamanAccurateClinicalBiomedical2022}, mainly rely on existing annotated datasets and therefore do not offer means of defining which entities to be extracted. 

As RadEx takes radiology reports as input, our framework is delimited from other systems that directly classify acquired images, as, for example, described by \citet{mckinneyInternationalEvaluationAI2020}, who developed a system that outperforms human readers in the prediction of breast cancer.  

Last, we want to emphasize the strengths of the proposed framework design: In contrast to other approaches as mentioned above, our information model allows clinicians to define facts to be extracted freely, as long as they adhere to the underlying information model constraints. For example, a fact might consist of only an unbroken span of text. Moreover, spans can overlap and still be determined due to the introduction of a specific anchor entity per fact. The decoupling of information extraction and template filling based on value standardizers reduces model complexity and allows for further modularization, enabling models for both tasks to be exchanged or improved independently. The information model is also convertible to a more commonly used model where entities and their relations are labeled separately.

\section{Conclusion}

In this paper, we introduced RadEx, a framework that support developing systems that clinicians allow the flexibility to define facts freely, provided they consist of an unbroken span of text. The framework decouples the task of information extraction from the  template filling task and reduces in this way model complexity and permits independent improvement of both tasks.

The RadEx framework supports development of clinical information systems. It offers a set of components and artifacts that can be implemented with concrete methods to realize an information extraction system as has been demonstrated in the case study. When for the single components concrete re-usable implementations are provided, this could speed up the development process. In this way, the framework facilitates the development of information extraction systems  by providing a standardized, re-usable infrastructure which helps avoiding that developers have to re-develop architectures from scratch. 

\section*{Ethics statement}

While RadEx is designed to facilitate the development of an information extraction system, the outputs should always be validated by experienced clinicians to avoid potential misinterpretations or omissions. Furthermore, automation of information extraction should not lead to a devaluation of human expertise.

We confirm that the development of the RadEx framework did not involve the use of any patient-related data. For the case study however, the usage of 210 de-identified mammography reports was approved by the ethics committee of the canton of Bern on June 16th, 2023 (ID: 2022-01621). 

Finally, we emphasize the importance of continuous stakeholder participation, especially from clinicians, to ensure that the system meets its intended objectives and that its adoption does not introduce new challenges or barriers in clinical workflows.

\section*{Conflict of Interest Statement}
The authors declare that the research was conducted in the absence of any commercial or financial relationships that could be construed as a potential conflict of interest.

\section*{Author Contributions}

DR: Conceptualization, Methodology, Software, Writing - original draft, Writing - review \& editing, Visualization; KD: Project administration, Supervision, Writing - review \& editing; JK: Conceptualization, Software; AS: Conceptualization, Writing - review \& editing

\section*{Funding}
This research was funded by the Swiss Innovation Agency, Innosuisse, grant number \textit{59228.1 IP-ICT}.

\section*{Data Availability Statement}
\label{sec:data}
Where mentioned in the main text, resources are available in an OSF repository, available at \href{https://doi.org/10.17605/OSF.IO/MNHXW}{10.17605/OSF.IO/MNHXW}.
The dataset generated and analyzed for this study is not publicly available due to institutional restrictions. However, additional details regarding the dataset are provided upon request. 

\bibliography{bibliography}

\end{document}